# A Bayesian Method for Causal Modeling and Discovery Under Selection

**Gregory F. Cooper**
Center for Biomedical Informatics
University of Pittsburgh
Pittsburgh, PA 15213
gfc@cbmi.upmc.edu

## Abstract

This paper describes a Bayesian method for learning causal networks using samples that were selected in a non-random manner from a population of interest. Examples of data obtained by non-random sampling include convenience samples and case-control data in which a fixed number of samples with and without some condition is collected; such data are not uncommon. The paper describes a method for combining data under selection with prior beliefs in order to derive a posterior probability for a model of the causal processes that are generating the data in the population of interest. The priors include beliefs about the nature of the non-random sampling procedure. Although exact application of the method would be computationally intractable for most realistic datasets, efficient special-case and approximation methods are discussed. Finally, the paper describes how to combine learning under selection with previous methods for learning from observational and experimental data that are obtained on random samples of the population of interest. The net result is a Bayesian methodology that supports causal modeling and discovery from a rich mixture of different types of data.

## 1 INTRODUCTION

Causal knowledge is central to science and to many other areas of inquiry. Experimental studies, such as randomized controlled trials (RCTs), often provide the most trustworthy methods we have for establishing causal relationships from data. Such studies, while potentially highly informative, may not be safe, ethical, logistically feasible, or financially worthwhile. Observational data are passively observed. Such data are more readily available than experimental data, and indeed, most databases are observational. Researchers have developed methods for causal modeling and discovery from observational data that are an unbiased sample from cases generated by a causal process of interest (Cooper and Herskovits 1992; Spirtes, et al. 1993; Heckerman, et al. 1995; Pearl 2000).

Not infrequently, however, observational data consists of a biased sample of the cases generated by the causal process of interest. The samples appear in a dataset due to some selection criteria or effect. Such a sample[1] is said to be subject to *selection bias*. As one example, a robot can only directly sample the terrain it can physically explore, which may not be representative of the entire terrain of interest. As another example, patients who come to an emergency room may not be representative (in all important ways) of patients in the entire population of interest[2]. Indeed, selection bias has been demonstrated empirically in several areas of medicine, as for example in (Gerber, et al. 1982). Nonetheless, we would like to use data collected in a given setting to induce causal relationships for that setting as well as for the broader population. This paper describes a method for modeling causal relationships under selection. Such modeling can be applied in performing causal discovery.

The concept of selection bias is well known (Sackett 1979). The idea of using a variable to represent selection is described in (Wermuth, et al. 1994). In (Spirtes, et al. 1993, Section 9.3) researchers discuss causal modeling from observational data when a population is sampled according to some particular selection criteria (e.g., all patients above a certain weight). Their approach distinguishes among causal models by using tests of conditional independence, rather than by using a Bayesian approach. In (Cooper 1995), numerous conditions under which causal structure and parameters can (and cannot) be learned from conditional-independence tests are described, when there is selection; a special-case Bayesian analysis of causal modeling under selection also is proposed. A general algorithm for causal discovery using conditional independence tests is developed in (Spirtes, et al. 1995). The unique contribution of the current paper is to describe a general Bayesian method for causal modeling and discovery under selection.

---

[1] Throughout the paper we use as synonyms the nouns *sample* and *selection*, the verbs *sample* and *select*, and the terms *sampled* and *selected*.

[2] In this paper, the term *population of interest* means a set of cases obtained by unbiased sampling from the cases generated by a causal process of interest. We use the term *total population* synonymously with *population of interest*.

## 2   BACKGROUND

A causal Bayesian network (or *causal network* for short) is a Bayesian network in which each arc is interpreted as a direct causal influence between a parent node (variable) and a child node, relative to the other nodes in the network (Pearl 1988). Figure 1 illustrates a hypothetical causal network structure, which contains five nodes. The probabilities (parameters) that are associated with this causal network structure are not shown.

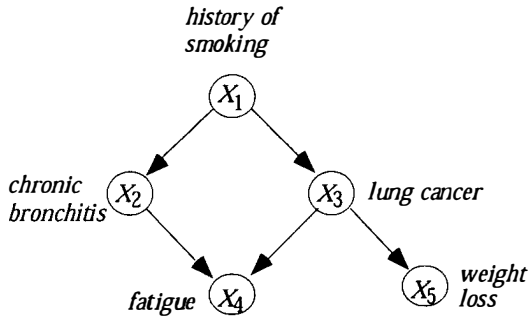

Figure 1: A hypothetical causal Bayesian network structure.

The causal network structure in Figure 1 indicates, for example, that a *history of smoking* can causally influence whether *lung cancer* is present, which in turn can causally influence whether a patient experiences *fatigue*.

The causal Markov condition gives the conditional independence relationships that are specified by a causal network:

> A node is independent of its non-descendants (i.e., non-effects) given its parents (i.e., its direct causes).

The causal Markov condition permits the joint distribution of the $n$ variables in a causal network to be factored as follows (Pearl 1988):

$$P(x_1, x_2, \ldots, x_n \mid K) = \prod_{i=1}^{n} P(x_i \mid \pi_i, K),$$

where $x_i$ denotes a state of variable $X_i$, $\pi_i$ denotes a joint state of the parents of $X_i$, and $K$ denotes background knowledge.

## 3   A BAYESIAN ANALYSIS

Researchers previously have described Bayesian approaches for deriving the posterior probability $P(M \mid D, K)$ of causal network structure $M$, given data $D$, subject to background knowledge $K$. Doing so requires (1) that a prior $P(M \mid K)$ on each possible causal network structure can be assigned, and (2) that a *marginal likelihood* $P(D \mid M, K)$ of the data given the model structure can be derived (Cooper and Herskovits 1992, Heckerman, et al. 1995). In the current paper, we focus on deriving $P(D \mid M, K)$ when $D$ contains data obtained under selection. While we concentrate on discrete variables and summation, the generalization of the concepts to continuous variables and integration will be obvious.

Due to space limitations, in this paper we do not describe how to learn the parameters (i.e., the probabilities) on $M$ given $D$ and $K$. However, given what is described here, in combination with previous literature on parameter learning in causal networks (Cooper and Herskovits 1992, Heckerman et al. 1995), the task is conceptually straightforward.

### 3.1  THE BASIC MODEL

A *case* denotes a set of values, one value for each variable in $M$. A case can be either measured (all values known), unmeasured (no values known), or partially measured (some values known). In this paper, we assume there is an underlying causal process that is generating cases that constitute the population of interest. We use $C$ to denote all the cases in the population of interest, regardless of whether those cases are measured, unmeasured, partially measured, or some combination thereof.

*Assumption* 1. The cases $C$ were generated by random sampling from the distribution of a causal network $B$ with structure $M$ and parameters $\theta_M$.

We will use $C_T$ to denote the sampled cases and $C_F$ to denote the unsampled cases. Set $C$ is the union of $C_T$ and $C_F$. Although by Assumption 1 we assume that $C$ was generated by random sampling from the distribution defined by $B$, in general this does not imply that each of $C_T$ and $C_F$ are themselves random samples from the distribution defined by $B$. In general, they will not be.

*Assumption* 2. Case selection is a causal event that can be modeled within a causal network that has a variable representing whether a case was selected.

To represent selection, we introduce a variable called $S$ into model $M$ that has states $T$ and $F$, which designate whether a given case was sampled ($T$) or not ($F$) (Cooper 1995). Let the term *model variables* designate all the variables in $M$, including $S$. Let the term *domain variables* denote all the variables in $M$, excluding $S$. We will use $n$ to denote the number of domain variables in $M$. In each case in $C_T$, variable $S$ has the value $T$, representing that the case was sampled. In each case in $C_F$, variable $S$ has the value $F$, denoting that it was not sampled. Thus, $S$ never has a missing value, because we know that a case either was or was not sampled.

*Example: Part* 1.

Table 1 shows an example dataset containing the five domain variables from Figure 1 and seven total cases that constitute the population of interest. The values of $S$ also are included. For this example, we might suppose (quite hypothetically) that there is a town with a total population of seven people, and that population count is known to us. Each of these people is a separate case. Three people in town have visited a *fatigue clinic*. It is the presence of fatigue that has caused these patient cases to appear as samples in the clinic's database. In Table 1, $S$ has the value $T$ for all three selected cases and the value $F$ for the four unselected cases. Also, variable $X_4$, which represents *fatigue*, has the value $T$ in all selected cases; in principle, however, $X_4$ might have had the value $F$ in some of the selected cases, because it is possible that the clinic would see a few people without fatigue. Most of the unselected cases do not have fatigue, but note that one does; in general, we do not expect that all cases in the population that are prone to being sampled will in fact be sampled.

Table 1: An example set of cases.

|        | selected cases | | | unselected cases | | | |
|--------|---|---|---|---|---|---|---|
| $X_1$  | $T$ | $F$ | $T$ | $T$ | $T$ | $F$ | $T$ |
| $X_2$  | $F$ | $T$ | $F$ | $F$ | $F$ | $T$ | $F$ |
| $X_3$  | $T$ | $F$ | $F$ | $T$ | $F$ | $F$ | $F$ |
| $X_4$  | $T$ | $T$ | $T$ | $T$ | $F$ | $F$ | $F$ |
| $X_5$  | $T$ | $F$ | $F$ | $T$ | $F$ | $F$ | $T$ |
| $S$    | $T$ | $T$ | $T$ | $F$ | $F$ | $F$ | $F$ |

☐

The parents of $S$ include the domain variables (e.g., fatigue) that are modeled as causally influencing whether a case is sampled. Let $\pi_S$ denote these variables. For the moment, we will assume that these variables are all measured, but later in this section we generalize to allow latent variables as well. We will include two additional variables as parents of $S$: variable $m_T$, which denotes the number of sampled cases, and variable $m_F$, which denotes the number of unsampled cases. These two variables will be considered part of model $M$. For now, we assume their values are known and are part of background knowledge $K$. The total number of cases in the population is then $m_T + m_F$. The reason for having $m_T$ and $m_F$ as parents of $S$ is that the probability that a case is selected from the total population will in general depend on the size of that population $(m_T + m_F)$ and the size of the sampled set $(m_T)$. For a given state of the domain parents $\pi_S$, as $m_T$ increases relative to $m_F$, typically $P(S = T \mid \pi_S, m_T, m_F, K)$ will increase, although in general the rate of increase will be sensitive to the value of $\pi_S$. In the limit of $m_T/(m_T + m_F) = 1$, $P(S = T \mid \pi_S, m_T, m_F, K) = 1$ for each state of $\pi_S$.

*Example: Part 2*

Continuing the example, Figure 2 shows a possible causal network structure to be evaluated using the known data in Table 1 and our prior beliefs.

As a prior for $S$ given its parents, we might, for example, use a Dirichlet distribution for which $P(S = T \mid X_4 = T, m_T = 3, m_F = 4, K) = 0.9$, $P(S = T \mid X_4 = F, m_T = 3, m_F = 4, K) = 0.01$, and the equivalent prior sample size is assumed to be 1 (Heckerman, et al. 1995).

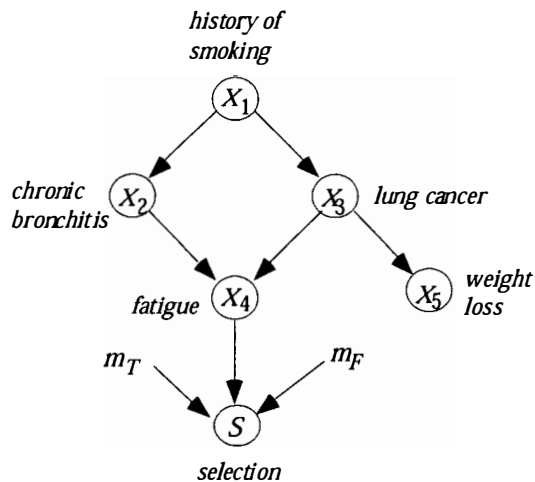

Figure 2: A modified version of Figure 1, which shows the explicit representation of selection using variable $S$. Variables $m_T$ and $m_F$ are explained in the text.

☐

Let $D_T$ denote the data that are known about the cases in $C_T$. Since for now we are assuming no missing values or latent variables, $D_T$ contains a value for each variable in each case in $C_T$. Let $D_F$ denote data that are known for the cases in $C_F$. Since these cases were not sampled, the only variable for which we know its value is $S$, which has the value $F$ for each case in $C_F$. Let $D$ denote the union of the known values in $D_T$ and $D_F$, which are all the data available to us.

*Example: Part 3*

We want to investigate the causal relationships among the five variables in the total population of seven cases. Table 2 shows the values that we know (as $T$ and $F$) and the values we do not know (as question marks). Set $D$ consists of the values that are known. Notice that for the unselected cases, the sole variable for which we know a value is variable $S$, because we know these four cases were indeed unselected.

Table 2: Data that are known when deriving the marginal likelihood.

|        | selected cases | | | unselected cases | | | |
|--------|---|---|---|---|---|---|---|
| $X_1$  | $T$ | $F$ | $T$ | ? | ? | ? | ? |
| $X_2$  | $F$ | $T$ | $F$ | ? | ? | ? | ? |
| $X_3$  | $T$ | $F$ | $F$ | ? | ? | ? | ? |
| $X_4$  | $T$ | $T$ | $T$ | ? | ? | ? | ? |
| $X_5$  | $T$ | $F$ | $F$ | ? | ? | ? | ? |
| $S$    | $T$ | $T$ | $T$ | $F$ | $F$ | $F$ | $F$ |

☐

As stated above, on the path to deriving $P(M \mid D, K)$, we derive $P(D \mid M, K)$. To do so, we will sum over the missing values of all the domain variables in all the cases in $C_F$; we use $H_F$ to denote the set over which we sum. The marginal likelihood of interest is therefore expressed as follows:

$$P(D \mid M,K) = \sum_{H_F} P(D, H_F \mid M,K). \qquad (1)$$

In the example, since there are 4 x 5 = 20 unknown variable values in the population of seven cases, Equation 1 will sum over the $2^{20}$ states of $H_F$.

The appeal of Equation 1 is that it expresses the marginal likelihood $P(D \mid M, K)$ on sampled cases in terms of a sum of marginal likelihoods on the total population of cases. This is useful because researchers previously have developed methods for deriving the marginal likelihood on the total population of cases, which is not subject to selection (Cooper and Herskovits 1992, Geiger and Heckerman 1994, Heckerman, et al. 1995).

In general, the size of the population will not be known with certainty, and thus, $m_F$ will be a random variable. Accordingly, we modify Equation 1 to sum over the possible values of $m_F$ as shown in Equation 2. In deriving $P(D \mid M, K)$, for fixed $D_T$ the parameter $m_T$ is a constant, and thus for notational simplicity we will assume that $m_T$ is part of background knowledge $K$.

$$P(D \mid M,K) =$$
$$\sum_{m_F} \sum_{H_F} P(D, H_F \mid M, m_F, K) P(m_F \mid M,K). \qquad (2)$$

Often, belief about $m_F$ might be independent of causal network structure $M$, and thus, the last term in Equation 2 would simplify to be $P(m_F \mid K)$.

We close this section with two relatively subtle, but important points. Set $C$ does not need to contain the entire population of interest (e.g., the entire county), but only an unbiased subset that includes the sampled cases (e.g., the town). Thus, if the town population is a random sample of the county population, defining $C$ as the town population is sufficient for deriving an unbiased marginal likelihood using Equation 2. Clearly less computation will be required when the number of unselected cases is fewer. The key condition is that $C$ be an unbiased sample from the distribution defined by the generating causal network of interest.

It may seem that if we are content to learn only about causal relationships for the sampled population, then we need not be concerned with modeling the unsampled population. Unfortunately, this is not so. The reason is that all causal network learning (of which we are aware) assumes a random sample from the joint distribution defined by $B$. The selected cases for which we have data are not a random sample. The following simple example illustrates the problem. Remove node $X_1$ from Figure 2, and consider the modified network to be $B'$, the generating network. In $B'$, $X_2$ and $X_3$ are marginally independent. If we condition on $S$ in $B'$, then in general $X_2$ and $X_3$ will be dependent, because they are d-connected (Pearl 1988). The selected population involves conditioning on $S = T$. Thus, in the selected population, $X_2$ and $X_3$ will likely be dependent, although they have no causal relationship between them *even in the selected population*. To estimate causal relationships for the selected population, we first need to estimate those relationships for the total population, then use that model to estimate the relationships for the selected population by setting $S = T$.

## 3.2 EXTENSIONS TO THE BASIC MODEL

As a generalization, suppose some modeled variables are latent and thus, they have no measurements, even in the selected cases. In the example, if $X_1$ were a latent variable, then the first row of Table 2 would contain all question marks. To derive the marginal likelihood of data $D$, we can modify Equation 2 to include an additional inner sum that sums over the values of the latent variables for the cases in $C_T$.

As another extension, consider multiple forms of case selection. If sets of cases were selected based on different criteria, then we simply need to create values for $S$ that designate which selection criterion was used for each set of cases. Equation 2 will remain applicable. The following example illustrates the basic idea.

*Example: Part 4*

Suppose that in the town of seven people there is also a *smoking clinic*, consisting of two people who are seeking help to stop smoking. Let $S = sc$ denote selection for these two cases. As before, let $S = fc$ designate selection for patients seen at the fatigue clinic. Finally, let $S = us$ represent the absence of selection for those people in the town who were not sampled. The modified dataset is shown in Table 3.[3]

Table 3: Data on three subpopulations, two of which were sampled.

|  | fatigue clinic cases | | | smoking clinic cases | | unsampled cases | |
|---|---|---|---|---|---|---|---|
| $X_1$ | $T$ | $F$ | $T$ | $T$ | $T$ | $?$ | $?$ |
| $X_2$ | $F$ | $T$ | $F$ | $F$ | $F$ | $?$ | $?$ |
| $X_3$ | $T$ | $F$ | $F$ | $T$ | $F$ | $?$ | $?$ |
| $X_4$ | $T$ | $T$ | $T$ | $T$ | $F$ | $?$ | $?$ |
| $X_5$ | $T$ | $F$ | $F$ | $T$ | $F$ | $?$ | $?$ |
| $S$ | $fc$ | $fc$ | $fc$ | $sc$ | $sc$ | $us$ | $us$ |

---

[3] For simplicity, in the example we assume that no one goes to both the fatigue clinic and the smoking clinic. To handle such a situation, we can create an additional selection value, namely $S = fc\_and\_sc$, to represent those cases that appear in both the subpopulations.

Figure 3 modifies Figure 2 to include the new influence of the smoking variable on selection. The prior distribution of $S$ given its parents also would be specified, but for brevity is not shown here.

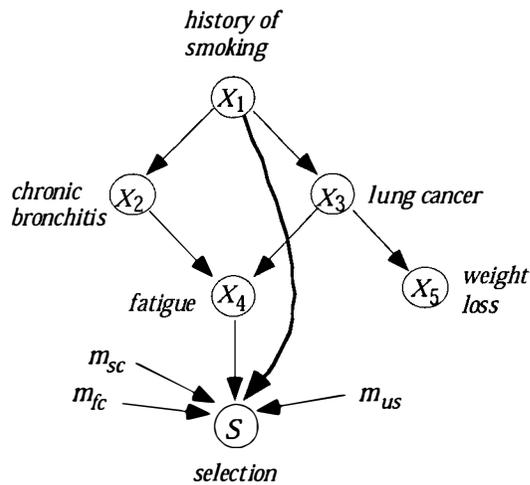

Figure 3: An extension of Figure 2 that models selection of two subpopulations.

To apply Equation 2, we replace $H_F$ in the equation with $H_{us}$.

□

So far, we have only discussed examples in which $S$ is a child of one or more domain variables. However, $S$ could also be a parent of some domain variables, if the selection event can causally influence the domain variables. For example, consider that the selected population consists just of people who visit the smoking clinic. Some of them may feel less fatigued than similar people in the entire population, because psychologically these smokers feel good about making an effort to improve their health by going to the smoking clinic. Thus, there would be an arc from *history of smoking* to $S$ and an arc from $S$ to *fatigue*.[4]

## 4   APPLICATIONS OF THE MODEL

In this section, we briefly describe how we can apply the method in Section 3 for causal modeling under selection when using a convenience sample and when using case-control data.

A *convenience sample* is a dataset that is collected because it is available, not necessarily because it is representative of the population of interest. Case-series reports in medicine are one type of convenience sample. The fatigue-clinic example in Section 3 involves a convenience sample. As another example, consider a survey that is distributed to a random sample of the population of interest. The people who complete and return the survey are a convenience sample. Arguably, most observational databases are convenience samples, at least to some extent. That is, few observational databases appear to represent a truly random sample of the entire population of cases that were generated by the causal process of interest. We can model with a convenience sample by using values $T$ (selected) and $F$ (unselected) for variable $S$, as in Section 3. For some modeling tasks, we may know the domain variables that influence selection. Thus, we need not search over models that contain different parents of $S$. In other situations, such search may be required; doing so would indicate the most likely causes of selection.

---

[4]   To handle the situation in which selection is also based on fatigue, we would need to add a temporal dimension to the model, in order to avoid directed cycles. We do not pursue that detail here, but only note that in modeling more complex, real-world events, temporal modeling will often come into play (see Section 6).

In case-control studies, which are common in medical research, an investigator identifies $m_1$ people with a given condition (the cases) and $m_2$ people without the condition (the controls). Often the condition is a disease, and the task is to discover the factors that causally contribute to having the disease. To model case-control studies, we can apply the method in Section 3 that involves using three values for $S$, namely the values *case*, *control*, and *unsampled (us)*. Let $m_3$ be the number of unsampled cases. The parents of $S$ are the domain variables that the investigator used as criteria to select the $m_1$ cases and the $m_2$ controls; the variables $m_1$, $m_2$, and $m_3$ are parents of $S$ as well.

## 5   COMPUTATIONAL ISSUES

In this section, we prove two complexity results. We then discuss some special-case and approximation methods.

***Theorem 1***   Deriving $P(D \mid M, K)$ under selection is NP-hard.
***Proof***   In (Cooper 1990), 3-SAT is reduced to causal network inference by using a network structure that includes a single node that has parents but no children. Let $S$ be that node. In the reduction, the inference task required to solve the 3-SAT problem is to derive $P(S = F)$. By assuming an empty set of selected cases (i.e., $m_T = 0$) and a single unselected case (i.e., $m_F = 1$), Equation 1 derives $P(S = F)$ for any model $M$, based on that model's prior probabilities. Thus, Equation 1 solves the 3-SAT problem.

□

***Theorem 2***   Finding the network structure $M$ that maximizes $P(D \mid M, K)$ under selection is NP-hard.
***Proof***   In the absence of selection, (Chickering 1996) showed that finding a network structure $M$ that maximizes $P(D \mid M, K)$ is NP-hard. For the same problem under selection, assume a structure prior in which there is zero probability that $S$ has any domain variables in $M$ as parents. Then Chickering's problem reduces to (indeed is equivalent to) finding the network structure $M$ that maximizes $P(D \mid M, K)$ under selection.

□

A proof parallel to the one in Theorem 2 can be used to reduce Bayesian network inference in the absence of selection (Cooper 1990) to Bayesian network inference under selection. Thus, for a given structure $M$ and database $D$, inference under selection also is NP-hard.

We now consider a method that can improve the efficiency of solving Equation 1. Let $A$ denote the set consisting of variable $S$ and the ancestors of $S$. Let $\overline{A}$ denote the nodes in $M$ that are not in $A$. Let $M^A$ designate the subgraph of $M$ on just the nodes $A$. Likewise, let $M^{\overline{A}}$ designate the subgraph of $M$ on just the nodes in $\overline{A}$. For the cases in $C_F$, let $H_F^A$ represent the state of the variables in $A$ that have missing values, and let $H_F^{\overline{A}}$ represent the state of the variables in $\overline{A}$ that have missing values. With this notation, we can rewrite Equation 1 as follows:

$$P(D \mid M,K) =$$
$$\sum_{H_F} P(D, H_F \mid M,K)$$
$$= \sum_{H_F^A} \sum_{H_F^{\overline{A}}} P(D, H_F^A, H_F^{\overline{A}} \mid M,K)$$
$$= \sum_{H_F^A} [\sum_{H_F^{\overline{A}}} P(H_F^{\overline{A}} \mid H_F^A, D, M, K)] P(D, H_F^A \mid M, K) \quad (3)$$

Since for the cases in $C_F$ we have by construction that $\overline{A}$ contains no nodes with fixed states, then the sum over the probabilities of the states of $H_F^{\overline{A}}$ is equal to 1. Thus the inner sum of Equation 3 (shown in square brackets) is 1 for any state of $H_F^A$ in the outer sum. Therefore, Equation 3 simplifies to be:

$$P(D \mid M,K) = \sum_{H_F^A} P(D, H_F^A \mid M,K). \quad (4)$$

Thus, to derive the marginal likelihood, for the unselected cases we need only sum over the states of the nodes that are ancestors of $S$. If $S$ were a root node in $M$, the sum in Equation 4 vanishes, and we simply compute $P(D \mid M, K)$ directly, without any need to model the unselected cases. Rarely will $S$ be a root note in $M$, however, because case selection is typically modeled as having a cause, and thus, $S$ will have parents.

Nevertheless, we could transform $M$ so that $S$ is a root node. Suppose we are using a class of causal network models for which Markov equivalence (aka independence equivalence) implies likelihood equivalence (Heckerman, et al. 1995)[5]. For example, the class of models that use multinomial distributions with Dirichlet priors satisfy these conditions (Heckerman, et al. 1995). Using such a class, suppose $S$ and its ancestors form a tree, that is, each internal node has one parent. We can apply Bayes rule to the prior parameters on $B$ in order to reverse all arcs away from $S$ (Shachter 1989), yielding a tree with the same connectivity, but no arcs in $S$. All trees with the same connectivity are Markov equivalent. We are assuming a model class in which Markov equivalence implies likelihood equivalence. By likelihood equivalence, the marginal likelihood for the transformed causal network will be the same as that of the original network. Figure 4a shows an example, where the cloud denotes an arbitrary causal subnetwork. In that figure, $S$ and its three ancestors form a tree.[6]

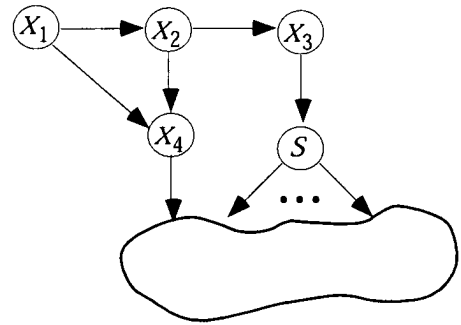

Figure 4a: The original network.

By reversing the arcs in that tree away from $S$ we create the network in Figure 4b that is Markov equivalent to the network in Figure 4a. Given the likelihood-equivalence assumption, the model in 4b will have the same marginal likelihood as the model in 4a for any dataset. Therefore, applying the line of reasoning above for when $S$ is a root node, we can use the network in Figure 4b to solve for $P(D \mid M, K)$ directly, without summation. We emphasize that this mathematical transformation does not change the semantics of the original network (Figure 4a) being scored, but rather, it merely scores that network efficiently.

---

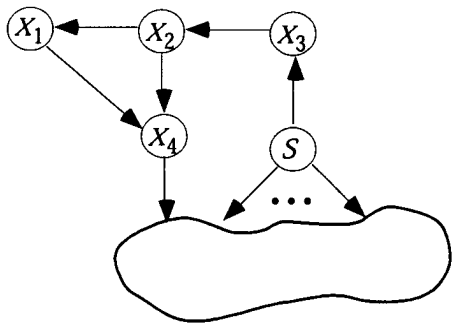

Figure 4b: The transformed network.

If $S$ and its ancestors do not form a tree, heuristically we could reverse arcs (Shachter 1989) away from $S$ anyway, thereby creating a network in which $S$ is a root node. We then can derive $P(D \mid M, K)$ without summation. There are, however, at least two problems with this approach. First, based on the line of reasoning in the proof of Theorem 1, it follows that the reversal task is NP-hard, and thus, quite likely intractable in the worst cases. Second, for non-tree-structured ancestors of $S$, the arc-reversal method will generate a Bayesian network that contains more parameters than the original network. Therefore, in general the marginal likelihood of the reversed network structure will not equal the marginal likelihood of the original network structure.

Figure 5 shows an example, where $M_1$ is the original network structure, and $M_2$ is the network structure after arc reversal. If the variables are binary and the probabilities (parameters) are represented using binomial distributions, then the joint probability distribution on $M_1$ is defined by 8 parameters, whereas the distribution on $M_2$ is defined by 10 parameters.

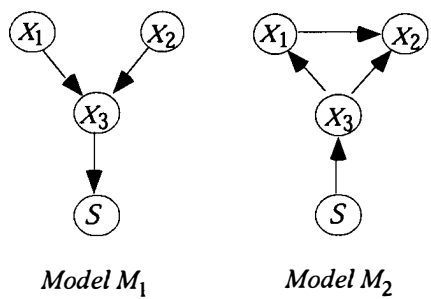

*Model $M_1$*              *Model $M_2$*

Figure 5: Original model $M_1$ and reversed model $M_2$, which contains arcs reversed away from $S$.

As a heuristic patch, we might approximate the marginal likelihood by using a measure like the BIC score (Geiger, et al. 1996) that contains a likelihood term that represents how well the model predicts the data and another term that represents the number of model parameters. The reversed model could be used to derive the likelihood term (e.g., $M_2$), whereas the parameter count would equal the number of parameters in the original model (e.g., $M_1$). It is an open problem to characterize and investigate how closely this heuristic score will approximate the correct marginal likelihood under different conditions.

## 6   LEARNING FROM A MIXTURE OF TYPES OF DATA

In this section, we describe how to model causal relationships when there are observational and experimental data, some of which are sampled under selection and some of which are sampled randomly from the population of interest.

Let $C_{ob}$ be a set of $m_{ob}$ cases that are randomly selected from the population of interest. In the general case, the event of random selection may itself causally influence some of the domain variables. We can create a value $ob$ for $S$ that indicates an observational case that was randomly selected.

We will represent manipulation of a variable $X$ by a variable $Q_X$, which has the same values as $X$, plus the value $ne$ that indicates a non-experimental case that is simply observed (see (Cooper and Yoo 1999) for details). $Q_X$ represents the value to which the experimenter intended to manipulate $X$. Ideally, $Q_X$ has no parents (and thus, its value is randomly set) and it deterministically controls $X$. That ideal may not be realized. As an example of imperfect control, a patient might initially agree to be in a study and to take a medication, but later refuse to take it reliably. If a case is not part of an experiment to manipulate $X$, then as stated, $Q_X = ne$. By introducing $Q_X$ variables, we transform learning with a mixture of experimental data and observational data into learning with observational data alone, since $Q_X$ is just another observation (Cooper and Yoo 1999).

In general, whether an experimental case appears in dataset $D$ may depend on the outcome of the experiment for that case. For example, patients who become ill from side-effects of a medication may leave the study unannounced. Thus, a model might contain an arc from a *drug side effects* variable to the variable $S$. We can create a value $ex$ for $S$ which indicates that data for the case (including outcomes) is recorded in the experiment's dataset.

*Example*
Figure 6 shows a causal network structure, which has a few changes from the network structures shown in Section 3. As in that section, this model represents a *hypothesis* about the underlying causal processes; it is not necessarily the generating network $B$. In Figure 6, variable $X_2$ (*current smoker*) is modeled as being experimentally manipulated in some cases. $X_4$ (*fatigue*) is modeled as the sole domain variable directly influencing case selection.

Table 4 contains a dataset used in deriving the marginal likelihood for the causal network in Figure 6 (as well as used in deriving the marginal likelihood of other possible causal network hypotheses that we wish to consider). Three of the cases in Table 4 are observational cases obtained under selection ($S = fc$), two cases involve an experimental manipulation ($S = ex$), two are observational cases that were randomly sampled ($S = ob$), and two cases remain unsampled ($S = us$).[7] Of course, a real dataset typically would have many more cases of each type. In this example, variables $X_1$ and $X_3$ are latent. Variable $X_2$ (*current smoker*) was experimentally manipulated in two cases.

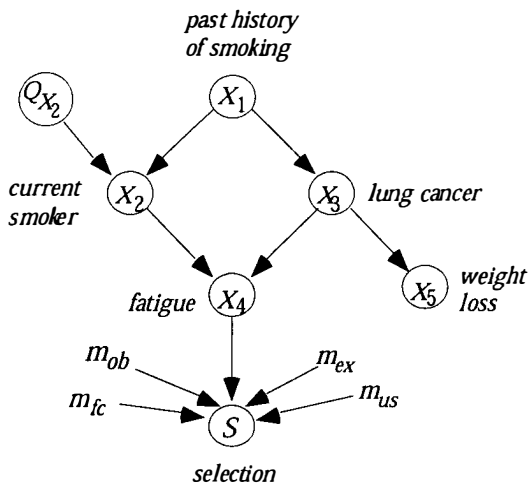

Figure 6: A causal network structure for modeling with observational and experimental data, some of which is sampled under selection and some of which is sampled randomly from the population of interest.

Table 4: Example of a mixture of cases that could be used for causal learning.

|  | fatigue clinic cases | | | smoking experimental cases | | randomly sampled observational cases | | unsampled cases | |
|---|---|---|---|---|---|---|---|---|---|
| $Q_{X_2}$ | ne | ne | ne | T | F | ne | ne | ne | ne |
| $X_1$ | ? | ? | ? | ? | ? | ? | ? | ? | ? |
| $X_2$ | F | T | F | T | F | T | F | ? | ? |
| $X_3$ | ? | ? | ? | ? | ? | ? | ? | ? | ? |
| $X_4$ | T | T | T | T | F | F | F | ? | ? |
| $X_5$ | T | F | F | T | F | F | F | ? | ? |
| $S$ | fc | fc | fc | ex | ex | ob | ob | us | us |

$Q_{X_2} = T$ means that the study participant was encouraged not to smoke, regardless of whether or not the participant had a history of smoking. $Q_{X_2} = F$ means that the experimenter made no attempt (one way or the other) to influence the participant's smoking behavior; thus, this experiment is only partially controlled. $Q_{X_2} = ne$ means that the case was not part of the experiment. In the two cases that are in the experiment, the value of $Q_{X_2}$ was set randomly, as represented in Figure 6 by $Q_{X_2}$ having no parents.

□

By using temporal causal networks (Dean and Kanazawa 1988) we can represent more complex types of data mixtures in which selection and experimentation occur over time. For example, at time $t_0$ a set of cases is selected according to some criteria (e.g., patients who arrive at a medical clinic during a given period). We represent this initial selection process as $S_{t_0} = medical\_clinic$. All those patients are asked to participate in an experimental study, and at time $t_1$ a subset agree to participate ($S_{t_1} = agree\_to\_participate$). At time $t_2$ the experiment is performed by randomly assigning subjects to either the experimental or the control treatment. At time $t_3$ a subset of the original subjects have remained in the study ($S_{t_3} = remained\_in\_study$) and their outcomes are measured and recorded. It is not uncommon in medicine for data to have a history as complex (or more complex) as the one in this example. The methods in this paper provide a basis for modeling both simple and complex forms of selection.

# 7  SUMMARY AND FUTURE WORK

This paper focuses on how to model causal processes using data that are obtained under selection. By developing a general model, it hopefully provides a useful foundation for further investigation. Key issues yet to be explored include the conditions under which causal network structure can be identified from data obtained under selection (possibly in combination with other types of data). Also, to attain computational tractability, it will be important to explore and characterize (both theoretically and empirically) approximations to the exact method described here. The usefulness of the model and its implementation will of course ultimately rest on how well they help us perform causal modeling and discovery with real data.

## Acknowledgments

The research reported here was supported by NSF grant IIS-9812021 and by NLM grant R01-LM06696. I thank Clark Glymour, Mehmet Kayaalp, Subramani Mani, Stefano Monti, Peter Spirtes, Changwon Yoo, and the UAI-2000 reviewers for helpful comments on earlier drafts of this paper. I also thank Constantin Aliferis for helpful discussions about causal modeling with temporal data obtained under selection.